\documentclass[conference]{IEEEtran}
\IEEEoverridecommandlockouts

\usepackage{cite}
\usepackage{graphicx}
\usepackage{placeins}
\usepackage{multirow}
\usepackage{amsmath,amssymb,amsfonts}
\usepackage{amsthm}
\usepackage{mathrsfs}
\usepackage[table]{xcolor}
\usepackage{textcomp}
\usepackage{booktabs}
\usepackage{etoolbox}
\patchcmd{\paragraph}{\itshape}{\bfseries}{}{\PackageError{paragraphstyle}{Could not set bold paragraph headings}{Check the document class paragraph definition.}}
\makeatletter
\patchcmd{\@makecaption}{\scshape}{\upshape}{}{\PackageError{tablecaptions}{Could not update table caption font}{Check the document class caption definition.}}
\makeatother

\usepackage{makecell}
\usepackage{tabularx}
\usepackage{siunitx}
\usepackage{soul}
\usepackage{url}
\usepackage[hypertexnames=false,hidelinks]{hyperref}

\usepackage{bm}

\title{Beyond Kinematic Tracking: Preserving Expressive Characteristics in Humanoid Whole-Body Control}

\author{Lingfan~Bao, Hanshang~Zhu, Yan~Pan, Jincheng~Wang, Jingyang~Yu,\\
Tianhu~Peng, Yiyang~Jia, Dimitrios~Kanoulas, and Chengxu~Zhou%
\thanks{This work was partially supported by the Advanced Research and
Invention Agency (grant number SMRB-SE01-P06) and the NVIDIA Academic
Grant Program.}%
\thanks{The authors are with the Department of Computer Science,
University College London, London, UK.
{\tt\small chengxu.zhou@ucl.ac.uk}}%
\thanks{Corresponding author: Chengxu Zhou.}%
}

\makeatletter
\IEEEaftertitletext{%
  \begin{minipage}{\textwidth}
    \centering
    \setlength{\abovecaptionskip}{3pt}
    \includegraphics[width=\textwidth]{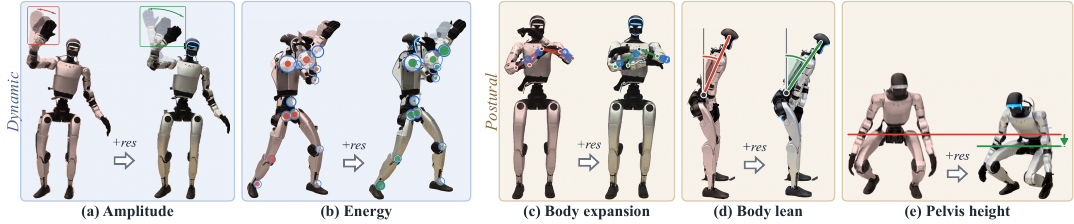}\par
    \def\@captype{figure}
    \caption{We demonstrate residual compensation for expressive motions on the Unitree G1. Pairs show executions without (left) and with (right) compensation for (a) amplitude, (b) energy, (c) body expansion, (d) body lean, and (e) pelvis height.}
    \label{fig:expressive_motion_teaser}
  \end{minipage}\par
}
\makeatother

\begin{document}
\bstctlcite{BSTcontrolThreeAuthors}
\flushbottom
\maketitle

\begin{abstract}
Whole-body control plays a crucial role in enabling humanoid robots to execute expressive motions. Faithful execution requires preserving both the intended action and its expressive characteristics. Existing whole-body trackers typically balance joint and keypoint tracking with physical stability across diverse motions, but these objectives do not necessarily prioritize the expressive characteristics of each action. Our analysis shows that lower average positional tracking error does not guarantee better preservation of dynamic and postural motion characteristics. To address this gap, we propose a residual learning framework with compensation objectives targeting amplitude, energy, body expansion, body lean, and pelvis height. To concentrate compensation while limiting interference with the underlying tracking behavior, motion-aware gating allocates correction capacity according to body-part activity, preserving larger correction ranges in more active parts and restricting them elsewhere. For each frozen tracker, we train motion-specific residual policies and distill them into a unified policy shared across motion categories. Simulation and physical experiments on the Unitree G1 across eight motion categories demonstrate superior preservation of dynamic and postural motion characteristics compared with the corresponding baselines.

\end{abstract}

\begin{IEEEkeywords}
Whole-Body Motion Planning and Control, Human and Humanoid Motion Analysis and Synthesis, Reinforcement Learning.
\end{IEEEkeywords}

\section{Introduction}\label{Sec:Introduction}

\begin{figure*}[!t]
    \centering
    \setlength{\abovecaptionskip}{2pt}
    \includegraphics[width=\textwidth,trim=38bp 148bp 42bp 82bp,clip]{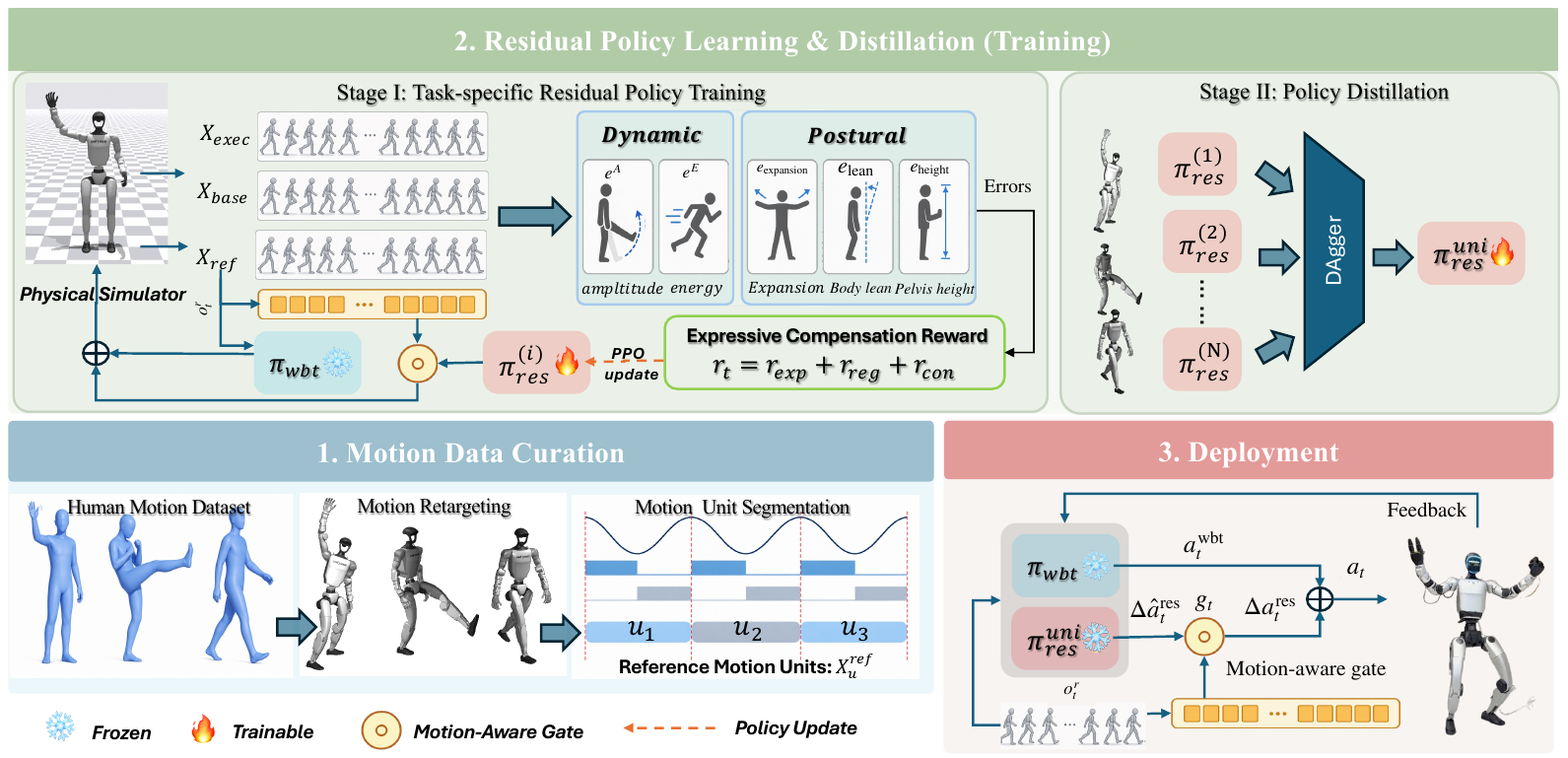}\par
    \caption{Residual compensation framework. (1) Human motions are retargeted, filtered, and segmented into units. (2) We train motion-specific teachers $\pi_{\mathrm{res}}^{(i)}$ (Stage I) and distill them into a unified student $\pi_{\mathrm{res}}^{\mathrm{uni}}$ (Stage II). (3) The student augments the frozen tracker $\pi_{\mathrm{wbt}}$, with motion-aware gating setting correction ranges by body-part activity.}
    \label{fig:expressive_tracking_pipeline}
\end{figure*}

Whole-body motion allows humanoid robots to convey intent and express themselves through physical behavior. Gestures and movements carry meaning through both the actions performed and how they unfold. Movement dynamics and posture are thus integral to the behaviors a robot seeks to reproduce. Bringing such expressiveness to physical humanoids requires whole-body control that coordinates movement while maintaining balance~\cite{xie2026textop,qi2026humanoidgpt,ZhengyiLuo_2026_SONIC,chen2025gmt,yanjie2025twist_corl,Chenxuxin2024expbody,ji2024exbody2}.

Recent advances in whole-body control have enabled humanoids to reproduce a broad repertoire of reference motions with a shared tracking policy~\cite{ZhengyiLuo_2026_SONIC,chen2025gmt}. Maintaining balance and satisfying physical constraints, however, may require deviations from the reference motion~\cite{ji2024exbody2}. While aggregate tracking errors measure whole-body motion fidelity, expressiveness may depend on particular postural and dynamic features, such as hand movement amplitude during waving or forward lean during bowing~\cite{bow,gross2012effort}. These features may be diminished despite low overall tracking error.

This raises two questions. \textit{How are expressive motion characteristics altered during physical execution?} \textit{How can we recover these characteristics while maintaining stable execution and accounting for tracking accuracy?}

Previous work has leveraged residual learning to improve tracking and adapt pretrained controllers to challenging motions~\cite{Kangning_Yin2026UniTracker}, object interaction~\cite{zhao2025resmimic}, and deployment differences~\cite{sun2026mosaic}. Residual policies refine execution by correcting nominal actions while retaining the underlying controller. This offers a way to address expressive characteristic discrepancies while keeping the pretrained tracker fixed.

To this end, we propose a residual learning framework for preserving expressive motion characteristics. Inspired by analyses of expressive movement~\cite{demeijer1989,gross2012effort}, we characterize dynamics through amplitude and energy, and posture through body expansion, body lean, and pelvis height. Discrepancies in these characteristics guide compensation by a unified residual policy shared across motion categories for each frozen tracker. To accommodate both postural and dynamic corrections, we introduce motion-aware gating that allocates larger correction bounds to more active body parts while retaining correction capacity in less active regions. Tracking constraints and bounded corrections further balance expressive preservation with tracking accuracy and stable execution.

Our contributions are threefold.
\begin{enumerate}
\item We analyze postural and dynamic deviations across motions and trackers, revealing that lower mean positional error does not necessarily correspond to better preservation of expressive motion characteristics.
\item We propose a motion-aware residual learning framework that augments a frozen whole-body tracker with a unified policy to recover expressive motion characteristics.
\item We conduct extensive evaluations across eight motion categories in simulation and on the physical Unitree G1. Results demonstrate that our framework outperforms the corresponding baselines in preserving dynamic and postural motion characteristics.
\end{enumerate}

\section{Related Work}\label{Sec:Related_work}

\subsection{Expressive Motion Analysis}

Human movement studies relate expressive qualities to movement dynamics and body posture, including trunk and arm configuration, movement direction, and intensity~\cite{demeijer1989,gross2012effort}. Research on bodily expression organizes these aspects into interpretable motion factors and quantitative measures~\cite{mahfoudi2023emotion_survey}. Laban-based studies associate combinations of movement elements with emotions elicited through movement~\cite{shafir2016emotion}, while computational approaches combine spatial body representations with energy-based temporal features for emotion recognition~\cite{WANG2025affectivebodyexpression}. Dynamic and postural features have also been examined in virtual avatars. Head inclination, head position, body openness, and joint velocities have been used to extract representative static poses from expressive walking, with perceptual evaluations supporting the transfer of expressive information~\cite{presti2022avatar}. Extending expressive movement to robots requires accounting for task requirements and physical embodiment. Work in this direction has explored learning expressive motion objectives from human feedback~\cite{9981964} and combining human motion style transfer with physics-aware generation and whole-body tracking~\cite{huang2026bionichumanmotionstyletransfer}. Drawing on these studies, we focus on the dynamic and postural characteristics that contribute to each motion's expressive qualities and use them to define explicit compensation targets for humanoid motion execution.

\subsection{Whole-Body Motion Tracking and Residual Adaptation}

Whole-body motion tracking enables humanoids to reproduce diverse reference motions while maintaining balance~\cite{bao2025deeploco,qi2026humanoidgpt,ZhengyiLuo_2026_SONIC,chen2025gmt,ze2026twist2,liao2025beyondmimic}. Expressive control has progressed from coordinating upper-body imitation with velocity-controlled locomotion~\cite{Chenxuxin2024expbody} to reproducing dynamic full-body motions~\cite{ji2024exbody2}. Recent residual and adapter-based methods further extend pretrained controllers to challenging motion distributions~\cite{Kangning_Yin2026UniTracker,wang2026FAST}, object interaction~\cite{zhao2025resmimic}, and changing task requirements~\cite{huang2026adamimic}. Related approaches address dynamics mismatch~\cite{HeT2025ASAP} and deployment-interface differences~\cite{sun2026mosaic}. In contrast to these adaptation objectives, our work focuses on preserving expressive motion characteristics by explicitly compensating for deviations in posture and movement dynamics.

\section{Method}\label{Sec:Overall_framework}

Figure~\ref{fig:expressive_tracking_pipeline} summarizes our framework, from motion curation to residual policy learning, distillation, and deployment.

\subsection{Problem Formulation}
We formulate expressive whole-body residual adaptation as goal-conditioned reinforcement learning (RL). The pretrained tracker $\pi_{\mathrm{wbt}}$ supplies nominal joint-position targets $\mathbf a_t^{\mathrm{wbt}}$. At step $t$, the residual policy observes $o_t=[o_t^p,o_t^r,e_t^{\mathrm{kp}}]$, comprising proprioception and control history, local target motion states, and keypoint tracking errors. Its output $\Delta\hat{\mathbf a}_t^{\mathrm{res}} =\pi_{\mathrm{res}}(o_t)\in[-1,1]^{n_a}$ yields $\mathbf a_t=\mathbf a_t^{\mathrm{wbt}} +\alpha\,\mathbf g_t\odot\Delta\hat{\mathbf a}_t^{\mathrm{res}}$, where $n_a$ is the action dimension of the underlying tracker, $\alpha=0.2\,\mathrm{rad}$, $\mathbf g_t\in[0.2,1]^{n_a}$ is the motion-aware gate, and $\odot$ denotes element-wise multiplication. We optimize the residual policy using proximal policy optimization (PPO)~\cite{schulman2017ppo} to maximize $\mathbb E_{\pi_{\mathrm{res}}}[\sum_{t=0}^{T-1}\gamma^t r_t]$, where $T$ is the horizon and $\gamma$ is the discount factor. Framewise and unit-completion rewards combine characteristic preservation, soft tracking constraints, and residual regularization.

\subsection{Data Curation}\label{sec:data_curation}
We select eight motion categories from AMASS~\cite{AMASS:ICCV:2019} using BABEL annotations~\cite{BABEL:CVPR:2021}: walk, kick, punch, crouch, wave, clap, cross arms, and bow, covering diverse limb movements and postural changes. After excluding unsupported interactions and mixed actions, we retarget the motions to G1 using SOMA~\cite{saito2026soma}, resample them to $50\,\mathrm{Hz}$, and apply PHC-based preprocessing~\cite{Luo2023PHC} to obtain robot reference trajectories.

To compute motion characteristics over complete movements or sustained poses, we segment references into units $u$ with trajectories $X_u^{\mathrm{ref}}$. Following the distinction between discrete and rhythmic movements~\cite{hogan2007rhythmic}, we identify one-shot movements, periodic cycles, and sustained holds. Limb-trajectory extrema, direction reversals, and inferred foot contacts guide movement boundaries; pelvis height, torso inclination, and relative hand positions identify posture changes and holds. Cycles span successive same-phase events, while one-shot units span onset to completion. We record each unit's type and frame boundaries according to its observed motion pattern, using the same intervals for characteristic computation and comparison across methods.

\subsection{Expressive Motion Descriptors}
\label{sec:descriptors}

\begin{figure}[t]
    \centering
    \setlength{\abovecaptionskip}{2pt}
    \includegraphics[width=0.6\columnwidth,trim=30bp 276bp 501bp 89bp,clip]{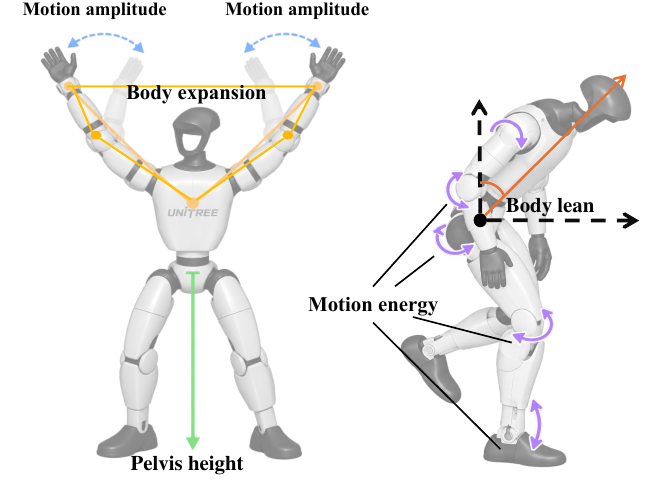}\par
    \caption{Kinematic cues for amplitude, energy, body expansion, body lean, and pelvis height.}
    \label{fig:descriptor_annotation}
\end{figure}

Following expressive movement analyses~\cite{gross2012effort,presti2022avatar}, we describe dynamics through amplitude and energy, and posture through body expansion, body lean, and pelvis height, as illustrated in Fig.~\ref{fig:descriptor_annotation}. We compute these descriptors over motion units of $T_u$ frames from $J=29$ joint angles and 14 forward-kinematic keypoints at $50\,\mathrm{Hz}$, with definitions given in Table~\ref{tab:motion_descriptors}. Throughout, $L$ and $R$ denote left and right. The five amplitude components are indexed by $\mathcal B=(\mathrm{LH},\mathrm{RH},\mathrm{LA},\mathrm{RA},\mathrm{W})$, corresponding to the left hand, right hand, left ankle, right ankle, and waist, respectively. The left and right shoulder origins are denoted by $\mathrm{LS}$ and $\mathrm{RS}$. For body expansion, $\mathcal S=\{L,R\}$ indexes the two sides, while $\mathcal K=\{e,w\}$ indexes the elbow and wrist points.

\begin{table}[!htbp]
\centering
\caption{Expressive motion descriptors computed over a motion unit $u$ of $T_u$ frames.}
\label{tab:motion_descriptors}
\small
\setlength{\tabcolsep}{2.5pt}
\setlength{\extrarowheight}{0pt}
\begin{tabular}{@{}lr@{${}={}$}lc@{}}
\toprule
Descriptor & \multicolumn{2}{c}{Definition} & Unit\\
\midrule
\addlinespace[2pt]
Amplitude & $\mathbf d_{\mathrm{amp},u}$ & $(a_{b,u})_{b\in\mathcal B}$ & m\\[-1pt]
 & $a_{\mathrm{W},u}$ & $a_{c,u}$ & \\[-1pt]
 & $a_{k,u}$ & $\displaystyle\max_{1\le t,t'\le T_u}\|\mathbf z_{k,t}-\mathbf z_{k,t'}\|_2$ & \\
\addlinespace[3pt]
Motion energy & $d_{\mathrm{energy},u}$ & $\displaystyle\frac{1}{T_u-1}\sum_{t=2}^{T_u}\sum_{j=1}^{J}(q_{t,j}-q_{t-1,j})^{2}$ & rad$^2$\\
\addlinespace[3pt]
Body expansion & $d_{\mathrm{exp},u}$ & $\displaystyle\frac{1}{T_u}\sum_{t=1}^{T_u}\sum_{s\in\mathcal S}\sum_{k\in\mathcal K}\|\mathbf p^{s}_{k,t}-\mathbf p_{c,t}\|_2$ & m\\
\addlinespace[3pt]
Body lean & $d_{\mathrm{lean},u}$ & $\displaystyle\frac{1}{T_u}\sum_{t=1}^{T_u}\sin\theta_t$ & --\\
\addlinespace[3pt]
Pelvis height & $d_{\mathrm{height},u}$ & $\displaystyle\frac{1}{T_u}\sum_{t=1}^{T_u}h_t$ & m\\
\bottomrule
\end{tabular}
\end{table}

\paragraph*{Dynamic descriptors}
Amplitude and energy describe movement extent and kinematic intensity~\cite{gross2012effort,presti2022avatar}. For reference and executed motions, we use pelvis-frame trajectories $\mathbf z_{k,t}=\mathbf R_{\mathrm{pelvis},t}^{\top}(\mathbf p_{k,t}-\mathbf p_{\mathrm{pelvis},t})$, where $\mathbf p$ is a world-frame position and $\mathbf R_{\mathrm{pelvis},t}$ maps pelvis to world coordinates. Hands use fixed fingertip offsets from wrist links; ankles use ankle-roll origins. The chest midpoint is $\mathbf p_{c,t}=(\mathbf p_{\mathrm{LS},t}+\mathbf p_{\mathrm{RS},t})/2$, whose trajectory diameter defines waist amplitude $a_{\mathrm{W},u}=a_{c,u}$. Motion energy uses squared increments of joint angles $q_{t,j}$ as a kinematic intensity proxy, not mechanical energy.

\paragraph*{Postural descriptors}
We describe body configuration through body expansion, body lean, and pelvis height, which remain informative during sustained holds~\cite{presti2022avatar}. Body expansion captures upper-body openness through the four elbow- and wrist-to-chest distances, $\ell^s_{k,t}=\|\mathbf p^s_{k,t}-\mathbf p_{c,t}\|_2$. Body lean characterizes torso inclination through $\sin\theta_t$, where $\theta_t$ is world-frame torso pitch, positive for forward bending. To capture body elevation, we measure pelvis height relative to the lowest sole point as $h_t=z_{\mathrm{pelvis},t}-\min_{s\in\mathcal S}z^s_{\mathrm{sole},t}$, where $z$ denotes world-frame height and $z^s_{\mathrm{sole},t}$ is the height of the lowest point on sole $s$.

\subsection{Motion-Aware Residual Control}
\label{sec:residual_compensation}
To accommodate both postural and dynamic corrections, the motion-aware gate assigns larger correction bounds to more active body parts while retaining correction capacity in less active regions. The gated residual is added to the nominal action as
\begin{equation}
\begin{aligned}
\Delta\mathbf{a}^{\mathrm{res}}_t &= \alpha\,\mathbf{g}_t\odot\Delta\hat{\mathbf{a}}^{\mathrm{res}}_t,\\
\mathbf{a}_t &= \mathbf{a}^{\mathrm{wbt}}_t+\Delta\mathbf{a}^{\mathrm{res}}_t.
\end{aligned}
\label{eq:residual_action}
\end{equation}

We divide the 29 joints into five body groups for motion-energy computation and gating. The left and right arms, denoted by $\mathrm{LH}$ and $\mathrm{RH}$, contain seven joints each; the left and right legs, denoted by $\mathrm{LA}$ and $\mathrm{RA}$, contain six each; and the waist $\mathrm{W}$ contains three. Group activity $r_{t,b}$ is measured by the root-mean-square reference joint velocity over up to ten future intervals. Let $r_t^{\max}=\max_b r_{t,b}$ and $b_j$ denote the group containing joint $j$. The gate is defined as
\begin{equation}
g_{t,j}=
\begin{cases}
0.2+0.8\,r_{t,b_j}/r_t^{\max}, & r_t^{\max}>0,\\
0.2, & r_t^{\max}=0.
\end{cases}
\label{eq:online_gate}
\end{equation}
For nonzero reference activity, the most active group receives the full correction bound of $\pm0.2\,\mathrm{rad}$, while all groups retain at least $\pm0.04\,\mathrm{rad}$ for postural adjustments. The gate uses only reference frames available to the actor. 

\begin{figure*}[!t]
    \centering
    \setlength{\abovecaptionskip}{2pt}
    \includegraphics[width=\textwidth]{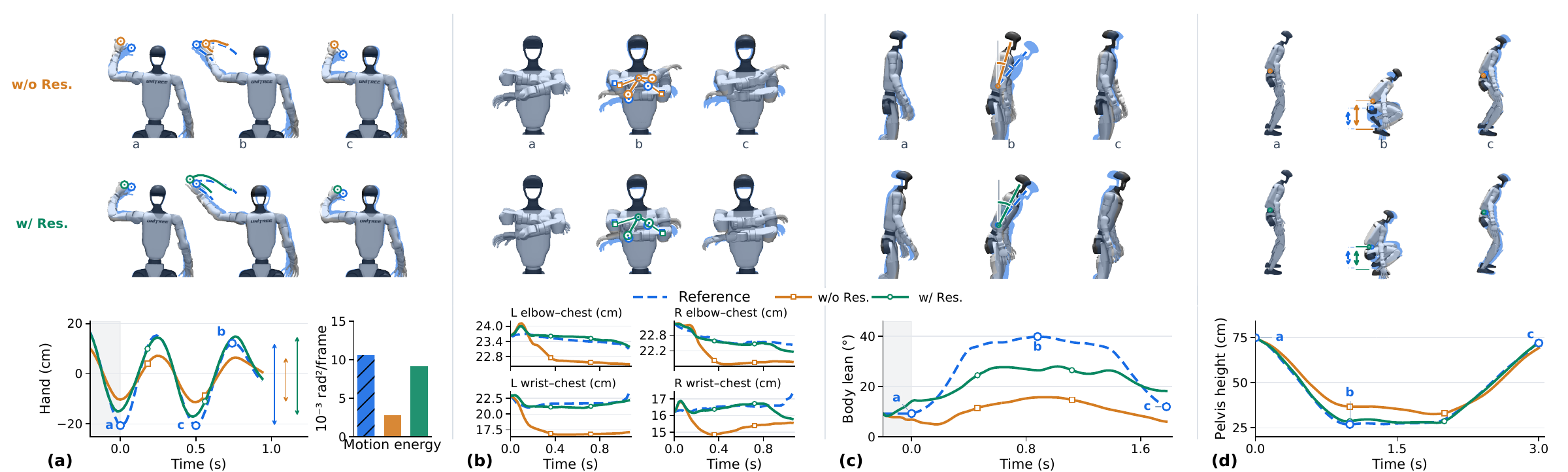}\par
    \caption{Compensation on representative development motions: (a) amplitude/energy, (b) upper-body geometry, (c) body lean, and (d) pelvis height. Blue ghosts show references; solid robots show execution without (w/o~Res.) and with (w/~Res.) compensation. Labels a--c link snapshots to times.}
    \label{fig:four_factor_comparison}
\end{figure*}

\subsection{Residual Policy Training}
\label{sec:policy_learning}

For each tracker, we train eight motion-specific residual policies $\pi_{\mathrm{res}}^{(i)}$, one per motion category, and distill them into a unified policy $\pi_{\mathrm{res}}^{\mathrm{uni}}$.

\noindent\textbf{State space design.} We define proprioception as $o_t^p=[\mathbf q_t,\dot{\mathbf q}_t,\mathbf b_t,\boldsymbol{\omega}_t,\mathbf a_{t-1}^{\mathrm{wbt}},\Delta\mathbf a_{t-1}^{\mathrm{res}}]$, where $\mathbf q_t$ and $\dot{\mathbf q}_t$ are joint positions and velocities, $\mathbf b_t$ and $\boldsymbol{\omega}_t$ are the gravity direction and angular velocity expressed in the body frame, and $\mathbf a_{t-1}^{\mathrm{wbt}}$ and $\Delta\mathbf a_{t-1}^{\mathrm{res}}$ are the previous nominal targets and applied residual corrections. The motion goal $o_t^r=[\mathbf q_{t-10:t+10}^{\mathrm{ref}},\mathbf b_t^{\mathrm{ref}}]$ contains a 21-frame reference joint-position window and gravity expressed in the current reference body frame, with indices clamped at sequence boundaries. Keypoint errors $e_t^{\mathrm{kp}}$ stack position differences (reference minus current) at elbows, knees, wrists, and ankles in their respective pelvis frames. The training critic additionally receives motion phase and its validity, base velocity, height, and foot contacts.

\noindent\textbf{Reward design.}
The reward $r=r_{\mathrm{exp}}+r_{\mathrm{con}}+r_{\mathrm{reg}}$ combines characteristic preservation, tracking constraints, and residual smoothness. We use $r_{\mathrm{exp}}=r_{\mathrm{amp}}+r_{\mathrm{energy}}+r_{\mathrm{expansion}}+r_{\mathrm{lean}}+r_{\mathrm{height}}$, with functions and weights in Table~\ref{tab:residual_rewards}.

Dynamic rewards average the five body groups equally. Amplitude uses $e^A_{b,u}=|a^{\mathrm{exec}}_{b,u}-a^{\mathrm{ref}}_{b,u}|$ at unit completion with a $0.02\,\mathrm{m}$ tolerance. Energy uses the absolute difference $e^E_{b,t}$ between executed and reference sums of squared joint-angle increments within group $b$ at consecutive frames.

Postural rewards compare body configurations at corresponding frames. We define $e^X_{s,k,t}=|\ell^{s,\mathrm{exec}}_{k,t}-\ell^{s,\mathrm{ref}}_{k,t}|$ for elbow/wrist-to-chest distances, $e^L_t=|\sin\theta^{\mathrm{exec}}_t-\sin\theta^{\mathrm{ref}}_t|$ for body lean, and $e^H_t=|h^{\mathrm{exec}}_t-h^{\mathrm{ref}}_t|$ for pelvis height. Amplitude and postural errors are mapped to bounded rewards using $\rho(z)=(1+z^2)^{-1}$ after normalization, with $[x]_+=\max(0,x)$ denoting the positive part.

Scales are shared across motion categories and trackers. We set $\sigma_A=0.04\,\mathrm{m}$, $\sigma_X=0.01\,\mathrm{m}$ for all four elbow/wrist-to-chest distances, $\sigma_{\mathrm{lean}}=0.04$, and $\sigma_{\mathrm{height}}=0.01\,\mathrm{m}$. Group energy scales $\sigma_{E,b}$ use reference motions pooled across the full training dataset. We calibrate $\lambda_E$ from tracker-only rollouts to balance amplitude rewards and energy penalties, then keep all scales and this weight fixed. Framewise postural and energy terms are divided by $T_u$ and $T_u-1$, respectively, to normalize unit duration; amplitude contributes once at completion.

\begin{table}[t]
\centering
\caption{Reward functions and weights for residual policy training.}
\label{tab:residual_rewards}
\small
\setlength{\tabcolsep}{4pt}
\renewcommand{\arraystretch}{1.45}
\begin{tabular}{@{}lcc@{}}
\toprule
Term & Function & Weight \\
\midrule
Amplitude
& $\displaystyle \frac{1}{|\mathcal B|}
\sum_{b\in\mathcal B}
\rho\!\left(\frac{[e^A_{b,u}-0.02]_+}{\sigma_A}\right)$
& $2$ \\

Energy
& $\displaystyle -\frac{1}{(T_u-1)|\mathcal B|}
\sum_{b\in\mathcal B}
\log\!\left(1+\frac{e^E_{b,t}}{\sigma_{E,b}}\right)$
& $\lambda_E$ \\

Body expansion
& $\displaystyle \frac{1}{4T_u}
\sum_{s\in\mathcal S}\sum_{k\in\mathcal K}
\rho\!\left(\frac{e^X_{s,k,t}}{\sigma_X}\right)$
& $1$ \\

Body lean
& $\displaystyle \frac{1}{T_u}
\rho\!\left(\frac{e^L_t}{\sigma_{\mathrm{lean}}}\right)$
& $1$ \\

Pelvis height
& $\displaystyle \frac{1}{T_u}
\rho\!\left(\frac{e^H_t}{\sigma_{\mathrm{height}}}\right)$
& $1$ \\
\midrule

Tracking constraint
& $-[v^{\mathrm{trk}}_u]_+$
& $\lambda_{\mathrm{trk}}$ \\

Residual action rate
& $\displaystyle -\frac{
\|\Delta\mathbf a_t^{\mathrm{res}}
-\Delta\mathbf a_{t-1}^{\mathrm{res}}\|_2^2
}{n_a\alpha^2}$
& $3$ \\
\bottomrule
\end{tabular}
\end{table}

\noindent\textbf{Soft constraints and regularization.}
The tracking constraint limits increases in positional error relative to the frozen tracker. For each motion unit $u$, let $m_u$ denote the mean wrist and ankle position error, $m_u^0$ its tracker-only value, and $\bar m^0$ the tracker-only mean across units. At unit completion, we apply $r_{\mathrm{con}}=-\lambda_{\mathrm{trk}}[v^{\mathrm{trk}}_u]_+$, where $v^{\mathrm{trk}}_u=(m_u-1.25m_u^0)/\bar m^0$ allows a $25\%$ increase before incurring a penalty. The multiplier $\lambda_{\mathrm{trk}}$ starts at zero and increases with the mean positive violation, up to a maximum of $10$.

Residual regularization $r_{\mathrm{reg}}$ penalizes squared changes in applied corrections at every control step without unit-length weighting, as defined in Table~\ref{tab:residual_rewards}.

\noindent\textbf{Initialization and termination.}
Episodes are initialized from reference poses and velocities at motion-unit boundaries or preceding run-ins. Run-ins provide tracking context before characteristic rewards become active. An episode ends when the reference motion is exhausted or a fall occurs, detected when pelvis height above ground drops below $0.15\,\mathrm{m}$. A fall replaces the final-step reward with $-1$. Incomplete units retain their framewise rewards but receive no unit-completion terms.

\noindent\textbf{Domain randomization.} The underlying trackers are pretrained with domain randomization for sim-to-real transfer~\cite{ZhengyiLuo_2026_SONIC,chen2025gmt}. We train residual policies under nominal dynamics.

\noindent\textbf{Policy distillation.} Following MOSAIC~\cite{sun2026mosaic}, we consolidate the motion-specific policies through dataset aggregation (DAgger). Teacher rollouts initialize the dataset, which is progressively augmented with student-visited states labeled by the corresponding teacher's clipped mean action. One unified policy per tracker minimizes action mean-squared error on the accumulated dataset, with equal sampling across motion categories. Deployment uses the same actor observations and motion-aware gating.

\section{Experiment} \label{Sec:results}
We evaluate our framework through three questions.
\begin{itemize}
    \item \textbf{Q1.} \textit{Do lower positional tracking errors imply better preservation of dynamic and postural characteristics?}
    \item \textbf{Q2.} \textit{How effectively does residual compensation preserve these characteristics, and how does it affect tracking accuracy and motion completion?}
    \item \textbf{Q3.} \textit{How does motion-aware gating affect residual intervention, tracking accuracy, and characteristic recovery?}
\end{itemize}

\subsection{Experimental Setup}
\label{subsec:Experiment_Setup}

We evaluate our framework in simulation and on the physical Unitree G1 across eight motion categories. We train residual policies in MuJoCo Warp~\cite{howell2025mujocowarp} with 1,024 parallel environments on NVIDIA RTX PRO 6000 GPUs. Using the AMASS motions curated in Sec.~\ref{sec:data_curation}, we compare SONIC~\cite{ZhengyiLuo_2026_SONIC}, GMT~\cite{chen2025gmt}, and TWIST2~\cite{ze2026twist2} with shared reference motions, unit boundaries, and evaluation starts. All trackers operate at $50\,\mathrm{Hz}$. Labels w/~Res. and w/o~Res. denote execution with and without residual compensation.

\paragraph*{Tracking baseline}
We include Track-Res, a tracking baseline that trains motion-specific residual policies to minimize mean pelvis-relative joint-position error. It uses two 256-unit hidden layers, as in our unified policy. We omit characteristic rewards but retain gating, regularization, and all constraints, including those on posture and body expansion.

\paragraph*{Metrics and protocol}
\label{sec:evaluation_metrics}
For each completed unit $u$, we measure errors in amplitude, energy, body expansion, body lean, and pelvis height:
\begin{equation}
\begin{aligned}
D_{A,u}&=\sum\nolimits_b
\left|a_{b,u}^{\mathrm{exec}}-a_{b,u}^{\mathrm{ref}}\right|,\\
D_{E,u}&=
\left|d_{\mathrm{energy},u}^{\mathrm{exec}}
-d_{\mathrm{energy},u}^{\mathrm{ref}}\right|,\\
D_{X,u}&=T_u^{-1}\sum\nolimits_{t,s,k}
\left|\ell_{k,t}^{s,\mathrm{exec}}
-\ell_{k,t}^{s,\mathrm{ref}}\right|,\\
D_{L,u}&=
\left|d_{\mathrm{lean},u}^{\mathrm{exec}}
-d_{\mathrm{lean},u}^{\mathrm{ref}}\right|,\\
D_{H,u}&=T_u^{-1}\sum\nolimits_t
\left|h_t^{\mathrm{exec}}-h_t^{\mathrm{ref}}\right|.
\end{aligned}
\label{eq:characteristic_errors}
\end{equation}
Here, $\mathrm{exec}$ and $\mathrm{ref}$ denote executed and reference motions. Sums span $t=1,\ldots,T_u$, $b\in\mathcal B$, $s\in\mathcal S$, and $k\in\mathcal K$.

\begin{figure*}[!t]
    \centering
    \setlength{\abovecaptionskip}{2pt}
    \includegraphics[width=\textwidth]{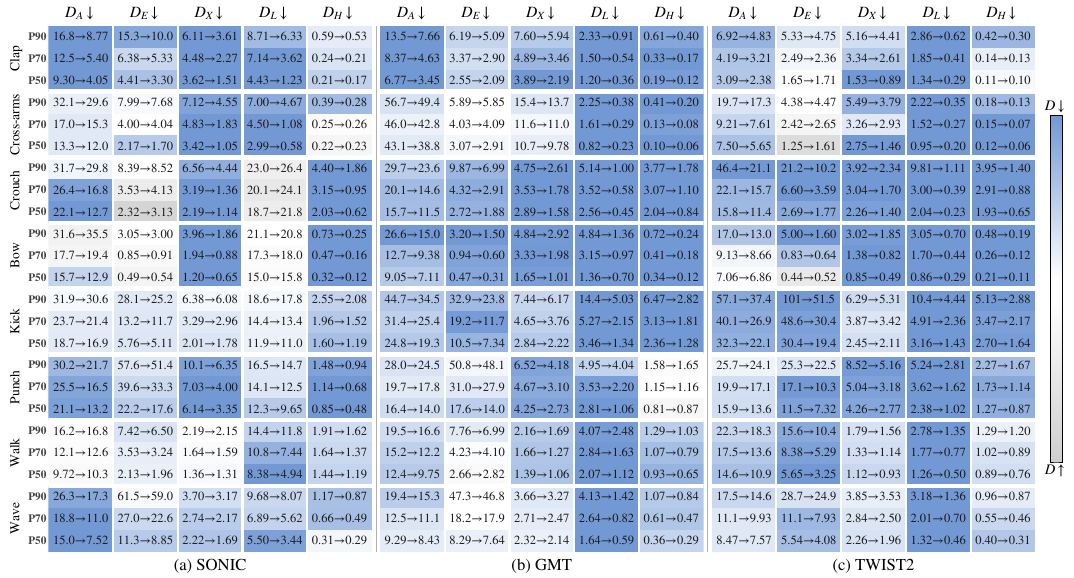}\par
    \caption{Characteristic errors for (a) SONIC, (b) GMT, and (c) TWIST2 at P90/P70/P50. Cells show w/o~Res. $\rightarrow$ w/~Res. using the units in Sec.~\ref{sec:evaluation_metrics}. Colors encode relative error reduction: blue indicates improvement, gray deterioration, and white changes within $\pm2\%$. Color saturation occurs at $\pm40\%$.}
    \label{fig:main_alignment}
\end{figure*}

For each tracker and action, we evaluate characteristic errors on units completed by all three methods in Table~\ref{tab:residual_objectives} or both conditions in Fig.~\ref{fig:main_alignment}, respectively. We use all evaluation starts for the main tracking and completion results. Tracking accuracy is measured by mean per-keypoint position error (MPKPE, cm), with pelvis-local mean per-joint position error (MPJPE) used for gate comparisons. We express $D_A$, $D_X$, and $D_H$ as percentages ($\%H$) of the fixed robot height $H_{\mathrm{G1}}=1.32\,\mathrm{m}$, and $D_E$ in $10^{-3}\,\mathrm{rad}^2$ per transition. Lean uses $(180/\pi)\arcsin(\min\{1,D_L\})$, a degree-equivalent sine discrepancy (deg-eq), not angular tracking error. P50, P70, and P90 denote percentiles. We weight the eight actions equally in all tables. Main tables average per-action P70 values from unit errors in simulation and reference-level mean errors in physical experiments. Arrows indicate preferred directions, and bold values identify the best result per tracker.

\begin{figure}[!t]
    \centering
    \setlength{\abovecaptionskip}{3pt}
    \includegraphics[width=\columnwidth]{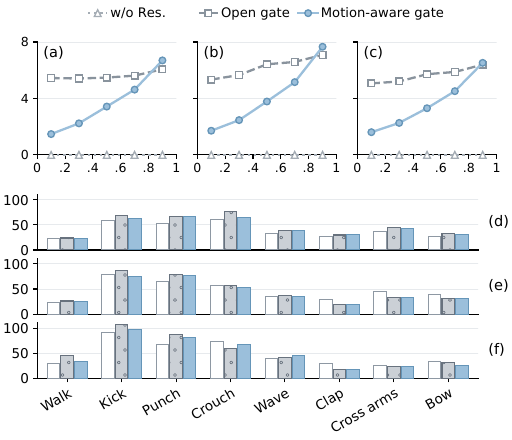}\par
    \caption{Motion-aware gating across SONIC, GMT, and TWIST2, in order within each group. (a)--(c) Applied correction (deg) versus normalized reference activity. (d)--(f) Low-activity limb endpoint error (mm; lower is better).}
    \label{fig:gate_combined}
\end{figure}

For trained-gate comparisons, we average available correction means over paired completed units within five equal bins of $r_{t,b}/r_t^{\max}$, then weight motions equally in Fig.~\ref{fig:gate_combined}(a)--(c). In (d)--(f), low-activity groups have duration-weighted unit root-mean-square (RMS) reference joint velocity below half the rollout maximum, fixed across policies; errors average eligible wrists/ankles, then rollouts. MPJPE uses common completed starts and equal motion and tracker weights.

\subsection{Expressive Preservation and Recovery}
\label{subsec:residual_compensation}

\begin{table}[!t]
\centering
\caption{We compare frozen tracking (w/o~Res.), tracking residuals (Track-Res), and expressive compensation (w/~Res.).}
\label{tab:residual_objectives}
\footnotesize
\setlength{\tabcolsep}{1.5pt}
\begin{tabular*}{\columnwidth}{@{\extracolsep{\fill}}llrrrrrr@{}}
\toprule
Tracker & Method & \makecell{MPKPE$\downarrow$\\(cm)} & \makecell{$D_A\downarrow$\\($\%H$)} & \makecell{$D_E\downarrow$\\($10^{-3}$ rad$^2$)} & \makecell{$D_X\downarrow$\\($\%H$)} & \makecell{$D_L\downarrow$\\(deg-eq)} & \makecell{$D_H\downarrow$\\($\%H$)} \\
\midrule
\multirow{3}{*}{SONIC} & w/o~Res. & 3.09 & 19.22 & 12.26 & 3.64 & 11.90 & 1.19 \\
 & Track-Res & \textbf{2.83} & 17.84 & 11.62 & 3.63 & 11.77 & 1.01 \\
 & w/~Res. & 3.30 & \textbf{14.79} & \textbf{10.66} & \textbf{2.13} & \textbf{10.72} & \textbf{0.71} \\
\midrule
\multirow{3}{*}{GMT} & w/o~Res. & 3.05 & 20.74 & 10.67 & 4.63 & 3.01 & 1.24 \\
 & Track-Res & \textbf{2.43} & 18.51 & 10.63 & 3.75 & 3.12 & 1.02 \\
 & w/~Res. & 3.04 & \textbf{17.22} & \textbf{9.01} & \textbf{3.60} & \textbf{1.15} & \textbf{0.72} \\
\midrule
\multirow{3}{*}{TWIST2} & w/o~Res. & 2.84 & 16.67 & 12.19 & 3.01 & 2.55 & 1.28 \\
 & Track-Res & \textbf{2.18} & 14.94 & 16.09 & 2.80 & 2.97 & 1.09 \\
 & w/~Res. & 2.91 & \textbf{12.84} & \textbf{7.90} & \textbf{2.29} & \textbf{0.87} & \textbf{0.73} \\
\bottomrule
\end{tabular*}
\end{table}

To answer Q1, we compare frozen tracking and Track-Res in Table~\ref{tab:residual_objectives}. We observe lower positional error with Track-Res for every backbone, but higher body lean error for GMT and higher energy and body lean errors for TWIST2. The tracker-only results in Fig.~\ref{fig:main_alignment} also show that characteristic errors vary across motions and backbones. These results show that lower average positional error does not ensure better preservation of every motion characteristic. Training budgets and distillation differ, so this comparison evaluates the resulting systems without isolating reward effects.

\begin{figure*}[!t]
    \centering
    \setlength{\abovecaptionskip}{2pt}
    \includegraphics[width=\textwidth]{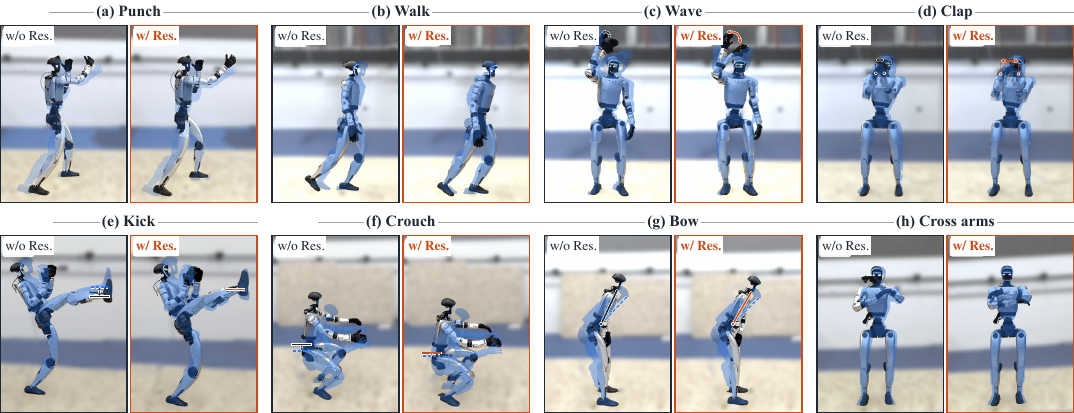}\par
    \caption{Physical executions without (left) and with (right) compensation; blue overlays show references. Arrows in (c) show hand excursion; markers show hand separation in (d) and projected reference deviations in (e)--(g).}
    \label{fig:real_world_robot}
\end{figure*}

To answer Q2, we evaluate expressive compensation in Table~\ref{tab:residual_objectives} and Fig.~\ref{fig:main_alignment}. We observe the lowest action-averaged errors in all five characteristics for each backbone. Positional error remains higher than Track-Res, while GMT slightly improves over frozen tracking. Across all trackers, the heatmap shows lower crouching expansion and height errors and lower waving amplitude errors at P50, P70, and P90.

We also examine where recovery remains incomplete. For SONIC crouching, amplitude error decreases but energy error increases at all reported percentiles. For TWIST2 walking, both errors decrease throughout. In the qualitative comparison in Fig.~\ref{fig:four_factor_comparison}, we observe improved pelvis height at the deepest crouch, while torso deviation persists.

We evaluate completion over all starts with equal action weights. Rates remain at $100\%$ for SONIC, change from $98.33\%$ to $98.27\%$ for GMT, and increase from $99.48\%$ to $99.89\%$ for TWIST2, remaining high with compensation.
\subsection{Ablation}
\label{sec:gate_ablation}

To answer Q3, we first compare motion-aware and open gates to examine correction allocation and tracking interference. Since opening the gate also changes the total correction range, we then compare against uniform gates at matched bounds to assess allocation with fixed policies at inference.

\noindent\textbf{Motion-aware versus open gates.}
We train both policies across all eight categories with identical dynamics, architecture, budget, objectives, constraints, regularization, and reference-activity observations, using $g_{t,j}=1$ for open gates. In Fig.~\ref{fig:gate_combined}(a)--(c), we observe smaller corrections at lower activity and comparable or larger corrections at the highest activity. Figure~\ref{fig:gate_combined}(d)--(f) shows lower low-activity endpoint errors for walking, kicking, crossing arms, and bowing across trackers, while effects on other motions vary. MPJPE decreases from $3.18$ to $3.10\,\mathrm{cm}$, compared with $2.97\,\mathrm{cm}$ for frozen tracking. These results show that gating concentrates compensation and reduces tracking interference in several motions.

\noindent\textbf{Motion-aware versus uniform gates.}
We replace motion-aware gates at inference with $g_t^{\mathrm{uni}}=(n_a^{-1}\sum_{j=1}^{n_a}g_{t,j}^2)^{1/2}$ on each controlled joint, matching squared correction bounds per frame. We keep the motion-aware-trained unified policies, observations, references, initial states, and evaluation seed fixed across all eight categories.

For Table~\ref{tab:uniform_gate}, group-energy error $D_E^{\mathrm{part}}$ sums absolute differences between unit-mean group energies. Low-limb error averages next-frame pelvis-local wrist/ankle errors for groups with reference activity at most one third of the frame maximum. Correction RMS measures applied corrections over controlled joints and valid unit frames. Results average common completed units with valid samples within each action.

\begin{table}[!t]
\centering
\caption{We compare uniform and motion-aware gates (MAG) at inference with matched correction-range budgets and fixed policies.}
\label{tab:uniform_gate}
\footnotesize
\setlength{\tabcolsep}{1.5pt}
\begin{tabular*}{\columnwidth}{@{\extracolsep{\fill}}llrrrrr@{}}
\toprule
Tracker & Gate & \makecell{$D_A\downarrow$\\($\%H$)} & \makecell{$D_E^{\mathrm{part}}\downarrow$\\($10^{-3}$ rad$^2$)} & \makecell{MPJPE$\downarrow$\\(cm)} & \makecell{Low limb$\downarrow$\\(cm)} & \makecell{Corr. RMS\\(deg)} \\
\midrule
\multirow{2}{*}{SONIC} & Uniform & 16.56 & 11.13 & 3.50 & 5.45 & 4.39 \\
 & MAG & \textbf{14.04} & \textbf{10.69} & \textbf{3.34} & \textbf{4.43} & 4.72 \\
\midrule
\multirow{2}{*}{GMT} & Uniform & 17.09 & 9.73 & 3.31 & 5.03 & 4.40 \\
 & MAG & \textbf{15.03} & \textbf{9.37} & \textbf{3.01} & \textbf{4.26} & 4.86 \\
\midrule
\multirow{2}{*}{TWIST2} & Uniform & 14.53 & 10.50 & 3.33 & 5.85 & 4.25 \\
 & MAG & \textbf{11.46} & \textbf{9.75} & \textbf{2.90} & \textbf{4.69} & 4.56 \\
\bottomrule
\end{tabular*}
\end{table}

In Table~\ref{tab:uniform_gate}, we observe lower amplitude, group-energy, MPJPE, and low-limb errors with motion-aware gating for every tracker. Equal-tracker reductions are $15.8\%$, $4.9\%$, $8.8\%$, and $18.1\%$, respectively, although walking energy error rises $3.3\%$. Applied correction RMS increases from $4.35^\circ$ to $4.72^\circ$ despite matched bounds. These results support activity-based allocation for the evaluated policies at inference, but do not establish superiority over uniform-gate training.

\subsection{Real-World Robot Experiments}
\label{subsec:real_world_experiments}

To answer Q2 on physical hardware, we evaluate residual compensation on the Unitree G1. We select three references per action and tracker across energy levels, screened in simulation for executability and expressive recovery. References vary across trackers but match between conditions. We run each reference once per condition, giving 48 executions per tracker and 144 in total. Errors use measured joint positions and pelvis orientation, with units averaged within references before aggregation in Sec.~\ref{sec:evaluation_metrics}.

We compare executions without and with compensation in Table~\ref{tab:real} and Fig.~\ref{fig:real_world_robot}. Lower errors in all five characteristics on each tracker's selected references indicate improved dynamic and postural preservation in these physical motions.

\begin{table}[!ht]
\centering
\caption{We compare executions without and with residual compensation on the physical Unitree G1.}
\label{tab:real}
\footnotesize
\setlength{\tabcolsep}{3pt}
\begin{tabular*}{\columnwidth}{@{\extracolsep{\fill}}llrrrrr@{}}
\toprule
Tracker & Method & \makecell{$D_A\downarrow$\\($\%H$)} & \makecell{$D_E\downarrow$\\($10^{-3}$ rad$^2$)} & \makecell{$D_X\downarrow$\\($\%H$)} & \makecell{$D_L\downarrow$\\(deg-eq)} & \makecell{$D_H\downarrow$\\($\%H$)} \\
\midrule
\multirow{2}{*}{SONIC} & w/o~Res. & 19.14 & 13.40 & 3.73 & 12.61 & 1.27 \\
 & w/~Res. & \textbf{15.73} & \textbf{11.46} & \textbf{2.97} & \textbf{11.93} & \textbf{1.05} \\
\midrule
\multirow{2}{*}{GMT} & w/o~Res. & 14.55 & 9.31 & 4.85 & 4.07 & 1.07 \\
 & w/~Res. & \textbf{10.53} & \textbf{8.09} & \textbf{4.39} & \textbf{2.54} & \textbf{0.91} \\
\midrule
\multirow{2}{*}{TWIST2} & w/o~Res. & 18.59 & 8.26 & 3.10 & 2.75 & 1.58 \\
 & w/~Res. & \textbf{16.98} & \textbf{7.21} & \textbf{2.96} & \textbf{2.28} & \textbf{1.15} \\
\bottomrule
\end{tabular*}
\end{table}

\FloatBarrier
\section{Limitations}
We use five descriptors to capture complementary aspects of expressive motion, but they do not cover all dynamic and postural characteristics. Agreement on these measures therefore does not imply preservation of every expressive cue. Unit-mean energy and lean errors can mask temporal discrepancies, and total energy can mask redistribution across body groups. Recovery also depends on the underlying tracker. SONIC can exhibit limited waist excursion despite retaining all three waist control channels. Adjusting body lean may then involve compensatory hip and leg motion. Such adjustments can alter limb amplitude, motion energy, and pelvis height, creating conflicts between recovery targets. Improving one characteristic may therefore worsen another.

\section{Conclusion}\label{Sec:conclusion}
In this work, we introduce a motion-aware residual learning framework for preserving expressive motion characteristics in humanoid robots. We use motion amplitude, motion energy, body expansion, body lean, and pelvis height as explicit compensation targets for a residual policy that augments a frozen whole-body tracker. Our comparisons show that lower mean positional tracking error alone is insufficient to ensure preservation of these characteristics. In the reported simulation comparison, residual compensation achieves lower aggregate errors in all five characteristics than tracking-oriented residuals, with higher positional error. Physical experiments with residual compensation on the Unitree G1 further demonstrate recovery of dynamic and postural characteristics.

\bibliographystyle{IEEEtran}
\bibliography{ral_conference}

\end{document}